\documentclass[runningheads,a4paper]{llncs}

\usepackage{graphicx} 
\usepackage{color}
\usepackage{times,helvet,courier}
\usepackage{latexsym}
\usepackage{xspace,epsfig,url,html}
\usepackage{epstopdf}

\def\ba{\begin{array}}
\def\ea{\end{array}}
\def\beq{\begin{equation}}
\def\eeq#1{\label{#1}\end{equation}}

\def\clasp{\htmladdnormallink{{\sc clasp}}{http://potassco.sourceforge.net/ }}
\def\dlv{\htmladdnormallink{{\sc dlv}}{http://www.dbai.tuwien.ac.at/proj/dlv}}
\def\dlvhex{\htmladdnormallink{{\sc dlvhex}}{http://www.kr.tuwien.ac.at/research/systems/dlvhex/}}
\def\gringo{\htmladdnormallink{{\sc gringo}}{http://potassco.sourceforge.net/ }}

\def\pharmgkb{\htmladdnormallink{{\sc PharmGKB}}{http://www.pharmgkb.org/ }}
\def\drugbank{\htmladdnormallink{{\sc DrugBank}}{http://redpoll.pharmacy.ualberta.ca/drugbank/ }}
\def\sider{\htmladdnormallink{{\sc Sider}}{http://sideeffects.embl.de/ }}
\def\biogrid{\htmladdnormallink{{\sc BioGrid}}{http://thebiogrid.org/ }}
\def\ctd{\htmladdnormallink{{\sc ctd}}{http://ctd.mdibl.org/ }}

\def\ape{\htmladdnormallink{{\sc ape}}{http://attempto.ifi.uzh.ch/site/ }}

\def\bqcnl{{\sc BioQueryCNL}}

\begin{document}

\mainmatter  

\title{Querying Biomedical Ontologies in Natural Language\\
using Answer Set Programming}

\titlerunning{Querying Biomedical Ontologies using ASP}

%
%
\author{Halit Erdogan\inst{1} \and Umut Oztok\inst{1} \and Yelda Erdem\inst{2} \and Esra Erdem\inst{1}}
\authorrunning{Erdogan, Oztok, Erdem and Erdem}

\institute{\vspace{-0.5ex}Faculty of Engineering and Natural
Sciences, Sabanc{\i} University, \.Istanbul, Turkey \and Research
and Development Department, Sanovel Pharmaceutical Inc., \.Istanbul,
Turkey }

%
%

\toctitle{Querying Biomedical Ontologies in Natural Language
using Answer Set Programming}

\tocauthor{Halit Erdogan, Umut Oztok, Esra Erdem}

\maketitle
%

Recent advances in health and life sciences have led to generation
of a large amount of data. To facilitate access to its desired
parts, such a big mass of data has been represented in structured
forms, like biomedical ontologies. On the other hand, representing
ontologies in a formal language, constructing them independently
from each other and storing them at different locations have brought
about many challenges for answering queries about the knowledge
represented in these ontologies.  One of the challenges for the
users is to be able represent a complex query in a natural language,
and get its answers in an understandable form: Currently, such
queries are answered by software systems in a formal language,
however, the majority of the users lack the necessary knowledge of a
formal query language to represent a query; moreover, none of these
systems can provide informative explanations about the answers.
Another challenge is to be able to answer complex queries that
require appropriate integration of relevant knowledge stored in
different places and in various forms.

\begin{figure}[t!]
    \centering
    \resizebox{2.5in}{!}{\includegraphics{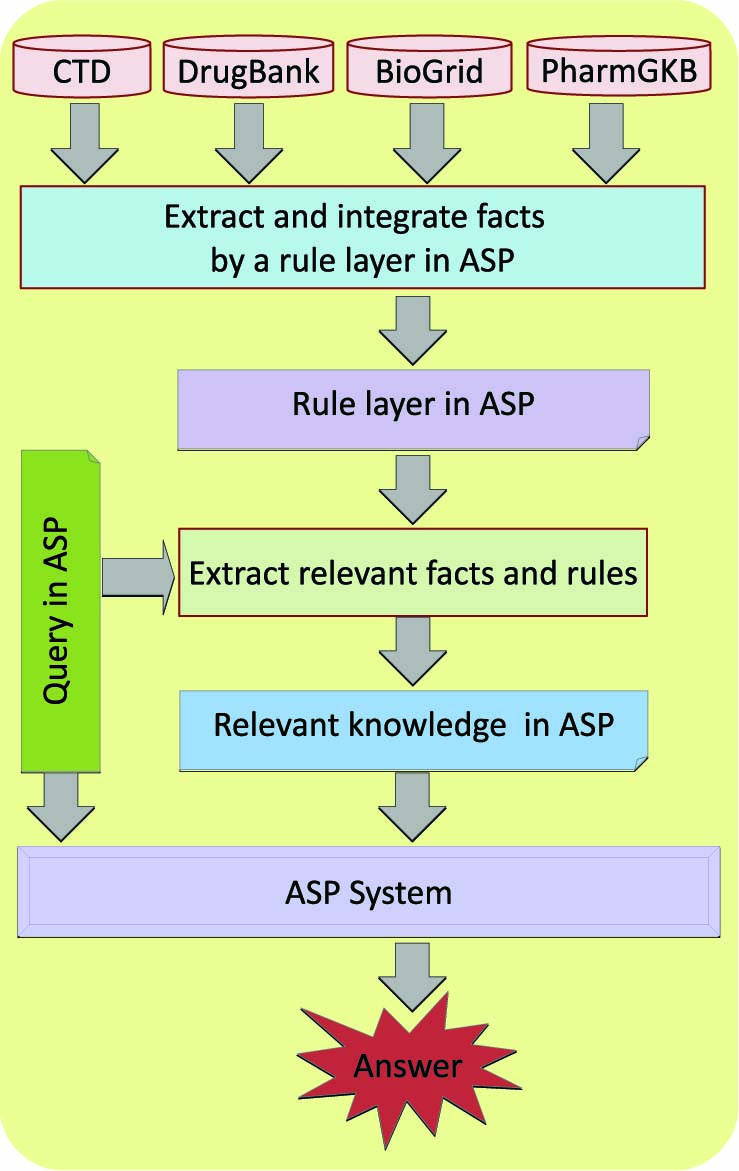}}
\caption{Extracting and integrating knowledge from ontologies or databases
that is relevant to a given query, and finding an answer to the query
using an ASP system.}
\label{fig:overall2}
\end{figure}

In this work, we address the first challenge by developing an intelligent
user interface that allows users to enter biomedical queries in a natural language, and
that presents the answers (possibly with explanations if requested)
in a natural language. We address the second challenge by developing
a rule layer over biomedical ontologies and databases,
and use automated reasoners to answer queries considering
relevant parts of the rule layer.
The main contributions of our work can be summarized as follows:

\begin{itemize}

\item
We introduce a controlled natural language, a subset of natural
language with a restricted grammar and vocabulary, specifically for
biomedical queries towards drug discovery; we call this controlled
natural language as \bqcnl~\cite{erdemYeniterzi09}.
For instance, in
this language, we can pose the following query:

\vspace{1ex}
\begin{quote}
``What are the
genes that are targeted by the drug Epinephrine and that interact
with the gene DLG4?"
\end{quote}
\vspace{1ex}

\begin{figure}[t!]
    \centering
    \resizebox{2.5in}{!}{\includegraphics{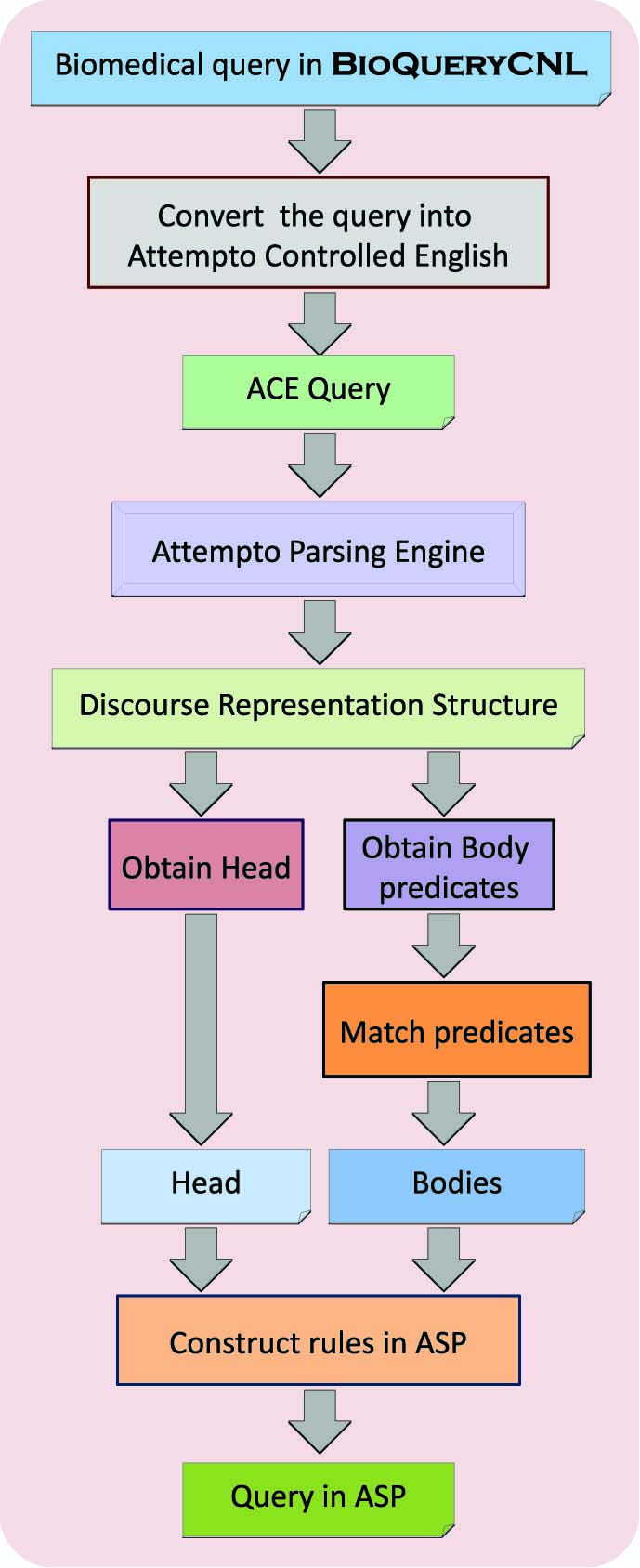}}
\caption{Transforming a query in \bqcnl\ into an ASP program.}
\label{fig:overall1}
\end{figure}

\item
We present an algorithm that converts  a biomedical query in \bqcnl\
into a program in answer set programming (ASP) --- a formal framework to automate reasoning about
knowledge~\cite{whatisasp-lifschitz-08} --- making use of the
parsing engine \ape~\cite{Fuchs00}. Figure~\ref{fig:overall1} shows
the overall idea behind this algorithm.
For instance, according to this algorithm, the query above is
translated into the following ASP program:
\begin{verbatim}
   what_be_genes(GN1) :-
      gene_gene(GN1,"DLG4"),
      drug_gene("Epinephrine",GN1).
\end{verbatim}
where {\tt gene\_gene} and {\tt drug\_gene} are defined in a ``rule layer''.

\item
Once we transform the biomedical query into an ASP program
and extract the relevant part of the rule layer (also an ASP program),
we can compute its answers (if exists) using a state-of-the-art ASP system,
such as \clasp~\cite{CLASPGebserSchaub-07},
\dlv~\cite{eiter97,leone06} or \dlvhex~\cite{dlvhex}, as described in \cite{bodenreiderCDE08}.
Figure~\ref{fig:overall2} shows the overall
idea behind this algorithm. For instance,  using \clasp, we compute the following answer
to the query above: ``ADRB1''.

\item
We construct an algorithm to provide minimal explanations to the
answers. For instance, for ``ADRB1'' our algorithm
provides the following minimal explanation:

\vspace{1ex}
\begin{quote}
the drug ``Epinephrine" targets the gene
``ADRB1" according to \ctd\ and the gene ``ADRB1" interacts with the gene ``DLG4'' according to
\biogrid.
\end{quote}

\end{itemize}

The applicability of our methods is illustrated with some complex
queries over \pharmgkb, \drugbank, \biogrid, \sider\ and \ctd, using
the ASP systems \clasp\ (with \gringo), \dlv\ and \dlvhex.

\subsubsection*{Acknowledgments.}
This work has been supported by TUBITAK Grant 108E229.

\bibliographystyle{splncs03}

\begin{thebibliography}{1}
\providecommand{\url}[1]{\texttt{#1}}
\providecommand{\urlprefix}{URL }

\bibitem{bodenreiderCDE08}
Bodenreider, O., Coban, Z.H., Doganay, M.C., Erdem, E.: A preliminary report on
  answering complex queries related to drug discovery using answer set
  programming. In: Proc. of the 3rd International Workshop on Applications of
  Logic Programming to the Semantic Web and Web Services (2008)

\bibitem{dlvhex}
Eiter, T., G.Ianni, R.Schindlauer, H.Tompits: Effective integration of
  declarative rules with external evaluations for {Semantic-Web} reasoning. In:
  Proc. of ESWC (2006)

\bibitem{eiter97}
Eiter, T., Leone, N., Mateis, C., Pfeifer, G., Scarcello, F.: A deductive
  system for non-monotonic reasoning. In: Proc. of LPNMR. pp. 364--375 (1997)

\bibitem{erdemYeniterzi09}
Erdem, E., Yeniterzi, R.: Transforming controlled natural language biomedical
  queries into answer set programs. In: Proc. of the Workshop on BioNLP. pp.
  117--124 (2009)

\bibitem{Fuchs00}
Fuchs, N.E.: Attempto controlled english. In: Proc. of WLP. pp. 211--218 (2000)

\bibitem{CLASPGebserSchaub-07}
Gebser, M., Kaufmann, B., Neumann, A., Schaub, T.: T.: Conflict-driven answer
  set solving. In: Proc. of IJCAI. pp. 386--392 (2007)

\bibitem{leone06}
Leone, N., Pfeifer, G., Faber, W., Eiter, T., Gottlob, G., Perri, S.,
  Scarcello, F.: The dlv system for knowledge representation and reasoning. ACM
  Trans. Comput. Log.  7(3),  499--562 (2006)

\bibitem{whatisasp-lifschitz-08}
Lifschitz, V.: What is answer set programming? In: Proc. of AAAI (2008)

\end{thebibliography}

\end{document}